\documentclass{article}
\usepackage{xurl}
\usepackage{array}
\usepackage{subcaption} 
\usepackage[english]{babel}
\usepackage{csquotes}
\usepackage{booktabs}
\usepackage{multirow}
\usepackage{adjustbox}
\usepackage{hhline}
\usepackage{makecell}
\usepackage{tabularx}
\usepackage{xcolor}   
\usepackage[letterpaper,top=2cm,bottom=2cm,left=2cm,right=2cm,marginparwidth=1.75cm]{geometry}

\usepackage[
  authoryear,        
  round,             
  sort&compress      
]{natbib}

\let\originalparencite\citep
\renewcommand{\citep}[1]{\textcolor{blue}{\originalparencite{#1}}}

\usepackage{amsmath}
\usepackage{amssymb}
\counterwithout{equation}{section}
\usepackage{parskip}
\usepackage{graphicx}

\title{ToolExpander: Extending the Frontiers of Tool-Using Reinforcement Learning to Weak LLMs}

\author{%
  \textbf{FuChen}$^{1,2}$\thanks{Work Done during an internship at OPPO.}\quad
  \textbf{PengWang}$^{1}$\quad
  \textbf{XiyinLi}$^{1}$\quad
  \textbf{WenLi}$^{1}$\quad
  \textbf{ShichiLei}$^{1}$\quad
  \textbf{DongdongXiang}$^{2}$\quad
  \\[2pt]
  $^{1}$OPPO \quad
  $^{2}$East China Normal University \\[1pt]
  \texttt{chenfu@stu.ecnu.edu.cn} \quad \texttt{wangpeng4@oppo.com}
}

\begin{document}

\maketitle

\begin{abstract}

Training Large Language Models (LLMs) with Group Relative Policy Optimization (GRPO) \citep{GRPO} encounters a significant challenge: models often fail to produce accurate responses, particularly in small-scale architectures. This limitation not only diminishes performance improvements and undermines the potential of GRPO but also frequently leads to mid-training collapse, adversely affecting stability and final efficacy. To address these issues, we propose \textbf{ToolExpander}, a novel framework that advances tool-oriented reinforcement learning for resource-constrained LLMs through two key innovations:
(1) \textbf{Dynamic Multi-Round Hard Sampling}, which dynamically substitutes challenging samples—those without correct outputs over 10 rollouts—with high-quality few-shot demonstrations during training, coupled with an exponential learning rate decay strategy to mitigate oscillations;
(2) \textbf{Self-Exemplifying Thinking}, an enhanced GRPO framework that eliminates KL divergence and incorporates adjusted clipping coefficients, encouraging models to autonomously generate and analyze few-shot examples via a minimal additional reward (0.01).
Experimental results demonstrate that ToolExpander significantly enhances tool-using capabilities in LLMs, especially in weaker small-scale models, improving both training stability and overall performance.
\end{abstract}

\section{Introduction}
In recent years, Large Language Models (LLMs) have demonstrated remarkable capabilities in natural language understanding, generation, and tool invocation. Reinforcement Learning (RL), particularly Reinforcement Learning from Human Feedback (RLHF) \citep{RLHF}, has emerged as a core methodology for model alignment and performance enhancement. As an efficient alternative, Group Relative Policy Optimization (GRPO) leverages ``inter-group relative rewards'' during policy updates, effectively mitigating issues such as sample redundancy and training instability prevalent in traditional RLHF. Consequently, GRPO has been widely adopted for tool invocation and decision optimization in Large Language Models.In our preliminary experiments, training that completely ignores hard samples leads to an extreme performance drop.

However, the practical application of GRPO faces significant challenges, especially when training small-scale models (e.g., at the 1.5B scale)\citep{Zhang2025NemotronResearchToolN1ET}. The limited capacity of these models leads to a high prevalence of ``hard samples''. The model repeatedly fails to generate correct answers for those hard samples. This abundance of hard samples not only drastically reduces the number of effective learning samples, thus limiting GRPO's potential, but also frequently triggers training collapse, where imbalanced parameter updates cause a sudden performance drop or even training failure. Furthermore, existing optimization strategies for GRPO often suffer from limitations: they either directly discard hard samples \citep{CPPO,DAPO}, simply increase the number of rollouts, or overcomplicate reward designs\citep{toolrl}. When the number of rollouts is directly increased from 10 to 32, the number of hard samples for the 1.5B model decreases by only 5\%–8\%, while the training time doubles. Models trained under complex reward signals may even underperform, for instance, the performance of 7B models can be surpassed by our 1.5B model. This phenomenon underscores the ineffectiveness of overcomplicated reward designs. 

To address the aforementioned challenges, this study aims to propose an enhanced GRPO reinforcement learning training approach. By incorporating few-shot guidance, we seek to reduce the number of hard samples in the training set and improve data utilization efficiency. We further demonstrate that this method is adaptable to both thinking and non-thinking modes of large language models. Moreover, we introduce a novel reasoning paradigm ``Self-Exemplifying Thinking", which enables the model to analyze and autonomously generate examples during the reasoning process.Based on this novel reasoning paradigm, we validate the generalizability of our strategy across both thinking and non-thinking modes.

Experimental results demonstrate that our scheme effectively addresses the core bottlenecks in small-model training. On two major tool invocation benchmarks, BFCL\citep{patilberkeley} and APIBank\citep{patilberkeley}, the number of hard samples for 1.5B and 7B parameter models decreased by an average of 15\%--20\%, with training stability significantly improved (no collapse occurred in 1.5B models). The optimized GRPO model substantially outperformed the original GRPO in standard inference mode. Notably, the Qwen2.5-7B model achieved an accuracy of 81.76\% on APIBank, surpassing both Tool-N1 (77.89\%) and the original GRPO (62.66\%). The ``Self-Exemplifying Thinking'' mechanism further enabled a secondary performance gain, allowing the model to maintain efficient decision-making on complex tasks without external guidance. Based on the core content of this paper, the framework is organized as follows:


\textbf{GRPO with few-shot Guidance} aims to reduce the proportion of hard samples in the training dataset through a dynamic multi-round hard sampling strategy.

\textbf{Self-exemplifying Thinking} guides the model to develop a cognitive framework capable of autonomously generating its own few-shot. 

Provoked by the data utilization efficiency achieved by few-shots learning in GRPO training, we propose a novel way of thinking. Specifically, self-exemplifying thinking enables the model to not only analyze few-shot instances but also generate such examples by itself when no pre-existing few-shot data is available.

\begin{figure}[t]  
    \centering  
    \includegraphics[width=0.6\textwidth]{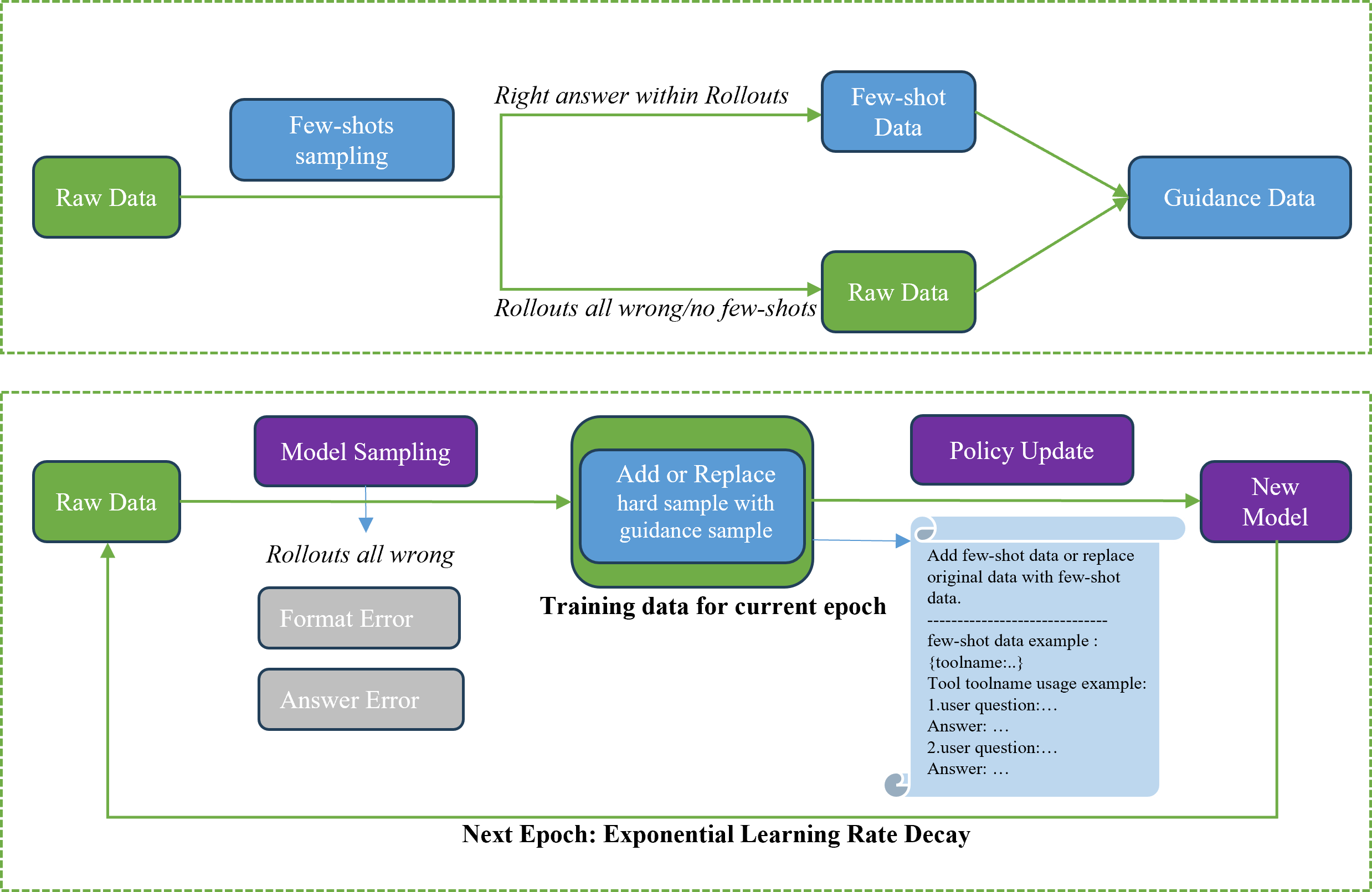}  
    \caption{The Overall Framework of ToolExpander}  
    \label{fig:overall framework}  
\end{figure}

\section{Related Work}
\textbf{Tool Calling.}
Early research in tool calling, exemplified by Toolformer \citep{ToolFormer}, successfully integrated tools for tasks such as question answering, calculation, translation, and calendar search. The subsequent popularization of function calling has granted Large Language Models (LLMs) access to an even broader range of tool resources. For large-scale tool calling scenarios, studies such as ToolLLM \citep{qin2023toolllm} have focused on guiding the model to plan tool usage by first filtering a relevant subset from a large tool pool. After tool filtering, enhancing the LLMs' capability for effective tool planning emerges as a key challenge. To address this, researchers have employed various post-training strategies. Methods like supervised fine-tuning (SFT) and Direct Preference Optimization (DPO)\citep{DPO} often exhibit a strong dependence on high-quality datasets. In contrast, reinforcement learning (RL) algorithms, including Proximal Policy Optimization (PPO)\citep{PPO} and Group Relative Policy Optimization(GRPO), place greater emphasis on enhancing the model's inherent reasoning abilities to improve tool planning.

\textbf{Reinforcement Tool Learning.}
Researches like Deepseek-R1 demonstrated that rule-based reinforcement learning (RL) can effectively enhance model reasoning. This paradigm is now gradually adopted in LLM-based tool calling. A study by ToolRL\citep{toolrl} empirically validated that the GRPO algorithm outperforms both PPO and SFT in tool calling tasks, highlighting the construction of complex reward functions as a key performance driver. Similarly, work on Tool-N1\citep{Zhang2025NemotronResearchToolN1ET} found that GRPO-based training yields superior results compared to supervised fine-tuning. However, a critical limitation was also identified: such training paradigms show limited effectiveness on small-scale models. This is primarily due to the unoptimized sampling process in standard GRPO, where small-scale models are frequently confronted with a large number of hard samples, leading to extremely low training data utilization efficiency.

These observations indicate that GRPO is not a flawless algorithm. To address its limitations, several improved variants have been proposed, such as DAPO\citep{DAPO} and GRPO-RoC\citep{rstar2}, which have achieved promising results. These works collectively underscore the critical importance of the sampling stage within GRPO. Team behind Deepseek-R1\citep{deepseekr1} proposed a complex multi-stage training strategy, using an initial simple RL training phase to ensure the model can generate valid results. In contrast, few-shot learning offers a more streamlined approach to enhance the model's ability to produce correct outputs.\citep{fewshots} Against this backdrop, we propose a method based on few-shot learning, specifically designed to reduce the prevalence of hard samples during the sampling process of the GRPO algorithm.

\section{GRPO with few-shot Guidance}


In this section, we leverage the GRPO algorithm to train the Qwen model for enhancing its function calling capabilities. During the training process, we directly adopt the official system prompts provided by Qwen without modifications. The core formulation of the GRPO algorithm is retained to ensure consistency with its original design intention. Our focus lies in investigating the performance differences between the base GRPO training and GRPO with few-shot guidance, particularly in terms of hard sample handling and overall performance improvement.

Formally, given the historical context \( c_t \) and the set of currently accessible tools \( \mathcal{Z} \), the Qwen model generates a set of candidate responses \( \{ O^1, O^2, \dots, O^N \} \), where each \( O^n \in \mathcal{O} \). Here, the \( \mathcal{O} \) denotes the space of possible output responses, and each response consists of two components: (1) textual reasoning tailored to tool-augmented problem-solving, and (2) an associated tool-related action \( a_n \). These candidate responses are evaluated by a reward function, producing a reward set \( \{ r_1, r_2, \dots, r_N \} \). We then optimize the policy \( \pi_\theta \) using GRPO, with the optimization objective formulated as:

\[
\begin{aligned}
\mathcal{L}_{\text{GRPO}}(\theta) =& \mathbb{E}_{(c_t, \mathcal{Z})} \mathbb{E}_{O^i \sim \mathcal{O}} \Bigg[ \min \left( \rho_i A_i, \text{clip}(\rho_i, 1 - \epsilon, 1 + \epsilon) A_i \right)\\ & - \beta \text{KL}\left( \pi_\theta \parallel \pi_{\text{old}} \right) \Bigg], \text{where } \rho_i = \frac{\pi_\theta(O^i \mid c_t, \mathcal{Z})}{\pi_{\text{old}}(O^i \mid c_t, \mathcal{Z})}
\end{aligned}
\tag{1}
\]

In this objective, \( \epsilon \) and \( \beta \) are tunable hyperparameters that balance the policy update range and the KL divergence constraint, respectively. \( A_i \) represents the relative advantage of the \( i \)-th response, which normalizes the raw reward \( r_i \) by the statistical properties of all candidate rewards in the batch:

\[
A_i = \frac{r_i - \text{mean}(\{ r_1, r_2, \dots, r_N \})}{\text{std}(\{ r_1, r_2, \dots, r_N \})}
\tag{2}
\]

where \( \text{mean} \) and \( \text{std} \) denote the mean and standard deviation of the rewards, ensuring that the advantage signal is scaled to guide policy updates effectively. In subsequent sections, we elaborate on three pivotal facets of our methodology: (1) the preparation and adaptive utilization of training data, encompassing the integration of Tool-N1 data, adaptation to Qwen's prompts, and construction of few-shot guidance dataset, (2) the design and implementation of dynamic multi-round hard sampling strategy to address training instability and sample utilization issues, and (3) the establishment of a reward mechanism.

\subsection{Few-shots construction}




In the early stages of training, we observed that for some small-scale models, the excessive number of hard samples makes it difficult to generate correct answers, resulting in very few samples that the model can learn through GRPO. Few-shot data can enhance the stability of small-scale models in the early training phase and convert some hard samples into learnable ones during the training process. Due to their limited capacity, small-scale models particularly 1.5B parameter series are highly prone to training failure or poor performance. We built a guidance dataset based on few-shot examples, which are sampled from the raw training dataset to solve the problem and support subsequent dynamic multi-round hard sampling training.

We adopted the training data from the researchers of Tool-N1 as the main dataset for our experiments. The dataset mainly consists of single-turn tool-using data obtained from xLAM\citep{zhang2024xlam} and ToolACE\citep{liu2024toolace} after preprocessing. Meanwhile, in a series of experiments, we adapted the experimental dataset to the official system prompts of Qwen. Unlike the researchers of Tool-N1, emphasizing that the model needs to conduct thinking-based reasoning in their prompts, the official system prompts of Qwen do not include such a requirement. After completing the series of experiments with Qwen official system prompts, we further aimed to enable the model to analyze examples through ``think'' and realize self-exemplification thinking. We first design novel system prompts tailored to the original training dataset. Concurrently, we conduct validation experiments to verify that, under the guidance of our proposed prompts, the integration of ``think" mechanisms and few-shot learning can synergistically further reduce the proportion of hard samples within the original training dataset.


In our experiments, we explored two approaches for constructing few-shot data: randomly constructed few-shot data and higher-quality few-shot data. The construction of random few-shot data involves extracting user queries and corresponding answers where each tool is utilized in other samples, and providing these as examples for the current sample. Importantly, the answer to the current sample is explicitly excluded from the few-shot examples to ensure more effective guidance for the model. For constructing higher-quality few-shot data, each selected example sample must ensure that the model can generate correct answers. If a currently selected example fails to enable the model to produce the correct answer, alternative examples are re-sampled until the correct answer is within rollouts. Otherwise, no few-shot data is added to that sample. 
\begin{table}[htbp]  
    \centering  
    \label{tab:basic}  
    \begin{tabular}{ccc}
        \hline
        Training data   & with few-shots & without few-shots \\  
        \hline
        62687   & 56113  & 6574   \\  
        \hline
    \end{tabular}
    \caption{ Data Statistics of Training Samples With Few-shot/Without Few-shot}  
\end{table}

\textbf{Details:} The random few-shot dataset also includes 6,574 samples without few-shot examples, as some tools lack other usage instances; these samples account for 10.49\% of the total data. Boosting rollouts to reduce hard samples is much less efficient than using few-shots.
Shown in Figure \ref{fig:rew_all} (c), the reward curve on the auxiliary few-shot dataset exceeds that on the original training dataset. However, training exclusively with few-shot data causes the model to over-rely on few-shot examples, leading to degraded inference performance on datasets without few-shot guidance.

\begin{figure}[htbp]  
    \centering  
    \includegraphics[width=0.5\textwidth]{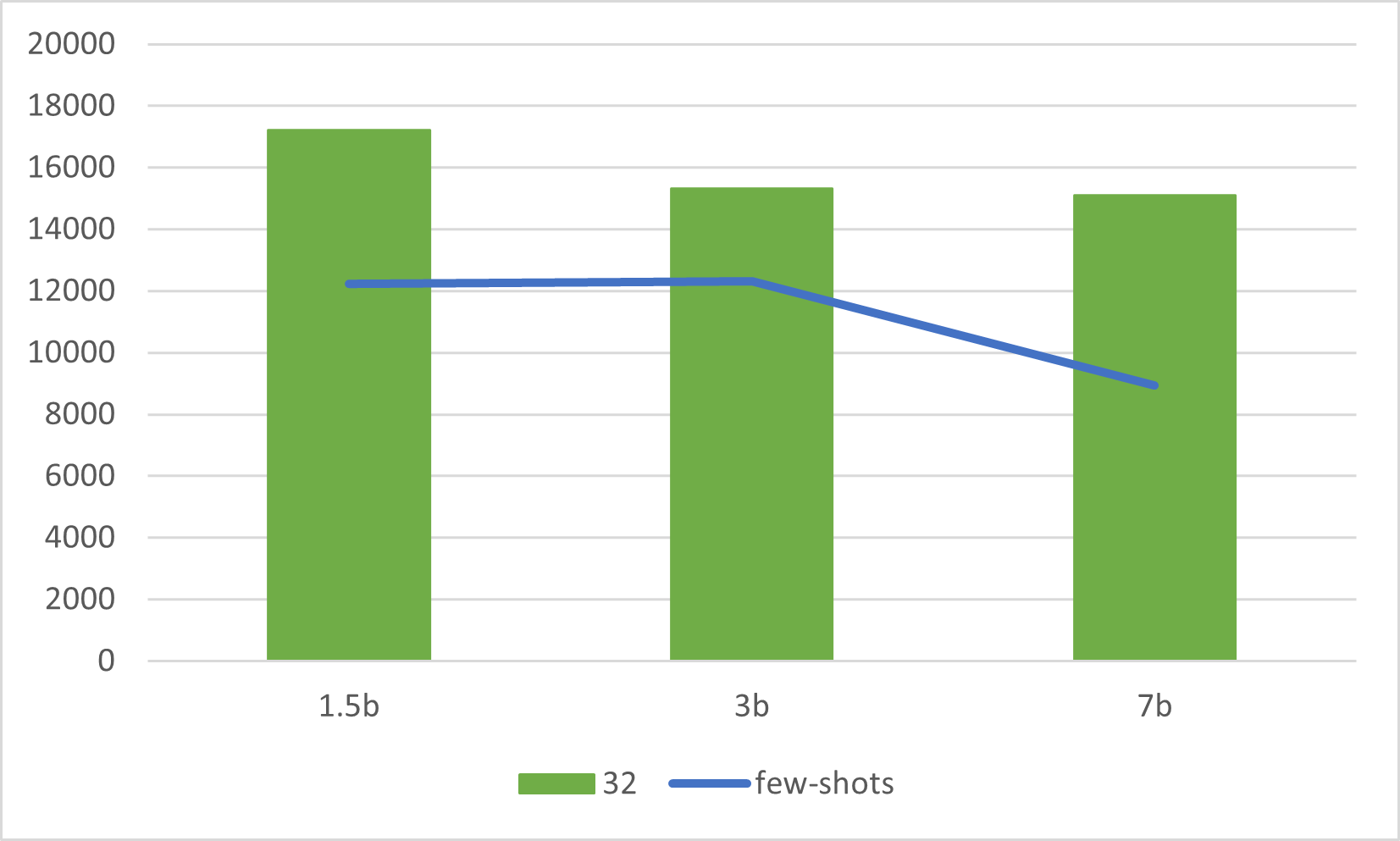}  
    \caption{Hard Samples Count: Few-shots vs. increasing the number of rollouts, rollout number is 10 for few-shots scenarios, rollout number is 32 for non-few-shots scenarios}  
    \label{fig:fsvsrollout}  
\end{figure}

\textbf{Analysis:} Few-shots help model generate correct answers. We did a sampling experiment to verify that few-shots reduce hard samples, shown in Table \ref{tab:hard_sample_count}. Notably, Qwen2.5-1.5B unexpectedly exhibits fewer hard samples than Qwen2-7B. This phenomenon suggests that the Qwen2.5 series are likely pre-trained on relevant training data. As mentioned in Tool-N1, the Qwen2.5-1.5B model shows limited potential for performance improvement when trained with GRPO, which may be attributed to its pre-training data coverage or inherent architectural characteristics that constrain further gains from this training paradigm. 

\begin{table}[htbp!]
    \centering
    \begin{tabular}{cccc} 
        \hline 
        Model & raw data & random few-shot data & better few-shot data \\ 
        \hline
        ToolN1-7b & 8527 & 8091 & 8044 \\
        Qwen2.5-1.5b-Inst & 17455 & 13611 & 12231 \\
        Qwen2-1.5b-Inst & 45362 & 38164 & 32381 \\
        Qwen2.5-7b-Inst & 15379 & 10507 & 8935 \\
        Qwen2-7b-Inst & 18168 & 10925 & 10520 \\
        \hline
    \end{tabular}
    \caption{Hard Sample Count. Note: The model cannot generate correct answers for hard samples within rollouts.} 
    \label{tab:hard_sample_count} 
\end{table}

\subsection{Dynamic multi-round hard sampling}


The entire training epochs are decomposed into multiple rounds. Before training, we did model inference for each question in raw data, shown in Figure \ref{fig:overall framework}. During each round, the hard samples of the current parameters are extracted, and these hard samples are either injected into our few-shot guided data or replaced with our few-shot guided data. Our experiments revealed that by adopting the dynamic hard sample replacement strategy, we achieved results on the BFCL test benchmark that are nearly comparable to those obtained with ``think" emphasized prompts. Tags ``<think> ... </think>" were not used in our experiments. Furthermore, on the APIBank benchmark, our approach significantly outperformed the training paradigm employed by Tool-N1.

\textbf{Reasons for Using the Dynamic Sampling Strategy}

During the training process of GRPO, the model is prone to encounter difficulties in sampling correct results. This issue is particularly acute for models with small parameter sizes, where the model struggles to generate accurate answers for a large number of samples, ultimately impeding effective training. To address this challenge, common approaches include constructing guided data to reduce the difficulty of the original data or excluding these problematic samples. However, training directly on guided data often leads the model to over-rely on such data, while excluding these samples significantly diminishes sample utilization. To resolve these issues, we have designed a strategy named Dynamic Multi-Round Hard Sampling.

We explored the use of entirely few-shot data for GRPO training and found that the model would over-rely on few-shot data. Specifically, the reward scores on the training set increased significantly, but the scores on the validation set without few-shot data were significantly lower than those obtained using the original GRPO training method.

Alternatively, we tried a two-stage training approach where few-shot data was used first, followed by training without few-shot data. It turned out that the performance of the model would gradually align with that of GRPO.

Mixing data in proportion also fails to yield satisfactory results. Simply excluding hard samples from the dataset leads to a reduction in data volume, which in turn causes original non-hard samples to be transformed into hard samples. Consequently, the performance of the trained parameters degrades significantly.

We observed that dynamically replacing hard samples with few-shot guided samples not only allows the model to gradually learn those samples that were previously difficult to master by leveraging the guided data while progressively breaking free from its reliance on such guided data but also significantly enhances sample utilization, reduces the number of hard samples, and simultaneously achieves performance improvements throughout the multi-round training process.

We did several experiments to analyze this phenomenon. Since small-scale models struggle to sample correct answers on the initial training dataset, it normally leads to training failures. We replace the data that still fail to generate answer correctly in 10 rollouts with few-shot data. This helps small-scale models learn from data easily. Moreover, in each round of training, sampling is still performed on the raw data, and all hard samples are replaced with few-shot samples that can correctly lead to the generation of correct results. Once the model generates a correct rollout for this sample, its few-shot data won't be involved anymore. During dynamic multi-round sampling, the number of hard samples gradually decreases, and the model gradually no longer relies on few-shot guided data. 

In the GRPO training process, for some hard samples that are beyond the model's capability boundary, the model cannot attain gradient updates. It fails to generate correct answer, leading to zero gradients. However, in our experiments, we found that the dynamic sampling strategy effectively addresses this issue. By replacing such hard samples with few-shot guided samples that can lead to correct results, it ensures that even samples that were originally beyond the model's capacity can contribute to gradient updates during training. 

\textbf{Strategy: add or replace}

In the dynamic sampling strategy, a key consideration is whether to add guided data or replace hard samples with guided data. We conducted discussions on whether to add hard sample data or replace hard sample data in dynamic multi-round sampling. Overall, the replacement approach outperforms the addition approach, and the training speed is faster. Whether hard samples are replaced with few-shot guided data or few-shot guidance is added, the number of hard samples in the original dataset ultimately decreases. 

Further analysis. Our strategy essentially involves injecting guided data. Changes in the dataset will inherently cause data fluctuations, and in principle, such fluctuations should be minimized. The injected data must be of high quality; otherwise, the model will tend to forget the information learned from the original data.

\begin{table}[htbp!]
    \centering
    \begin{tabular}{cccc} 
        \hline 
        Model & GRPO & Dynamic replace & Dynamic add \\ 
        \hline
        Qwen2.5-1.5b-inst & 12211 & 10458 & 11570  \\
        Qwen2-1.5b-inst & 14211 & 12488 & 13487 \\
        Qwen2.5-7b-inst & 9333 & 9157 & 9223  \\
        Qwen2-7b-inst & 10488 & 10347 & 10475 \\
        \hline
    \end{tabular}
    \caption{The number of hard samples of the model on the original dataset: the dataset uses Qwen's official system prompts without involving the model's ``think'' process, and the guided data adopts random few-shot.The 1.5B model experiences training collapse when using the original GRPO training method.} 
    \label{tab:number of hard samples} 
\end{table}

\begin{table}[htbp!]
    \centering
    \begin{tabular}{cccc} 
        \hline 
           & GRPO & More Cautious & More Bold \\ 
        \hline
        hard samples & 9333 & 9150 & 9811 \\
        BFCL ACC    & 0.8452 & 0.8511 & 0.8477 \\
        \hline
    \end{tabular}
    \caption{Bold or Cautious: ``More Bold'' means that some few-shot examples fail to guide the model to generate correct answers, while ``More Cautious'' ensures that only those few-shot examples capable of guiding the base model to produce correct answers are used. In this set of comparative experiments, the training parameters and training epochs are kept consistent, with Qwen2.5-7b-instruct employed.} 
    \label{tab:bold_or_catious} 
\end{table}

\textbf{Exponential Learning Rate Decay}


During the training process that utilizes a dynamic sampling strategy, the interchange between the raw training data and the auxiliary guidance data leads to different training data in each round. Although this variability increases the amount of learning information, it also exacerbates training-induced forgetting. To reduce the oscillation in the number of hard samples caused by forgetting phenomenon, we implemented exponential learning rate decay.

\begin{figure}[htbp]  
    \centering  
    \includegraphics[width=0.5\textwidth]{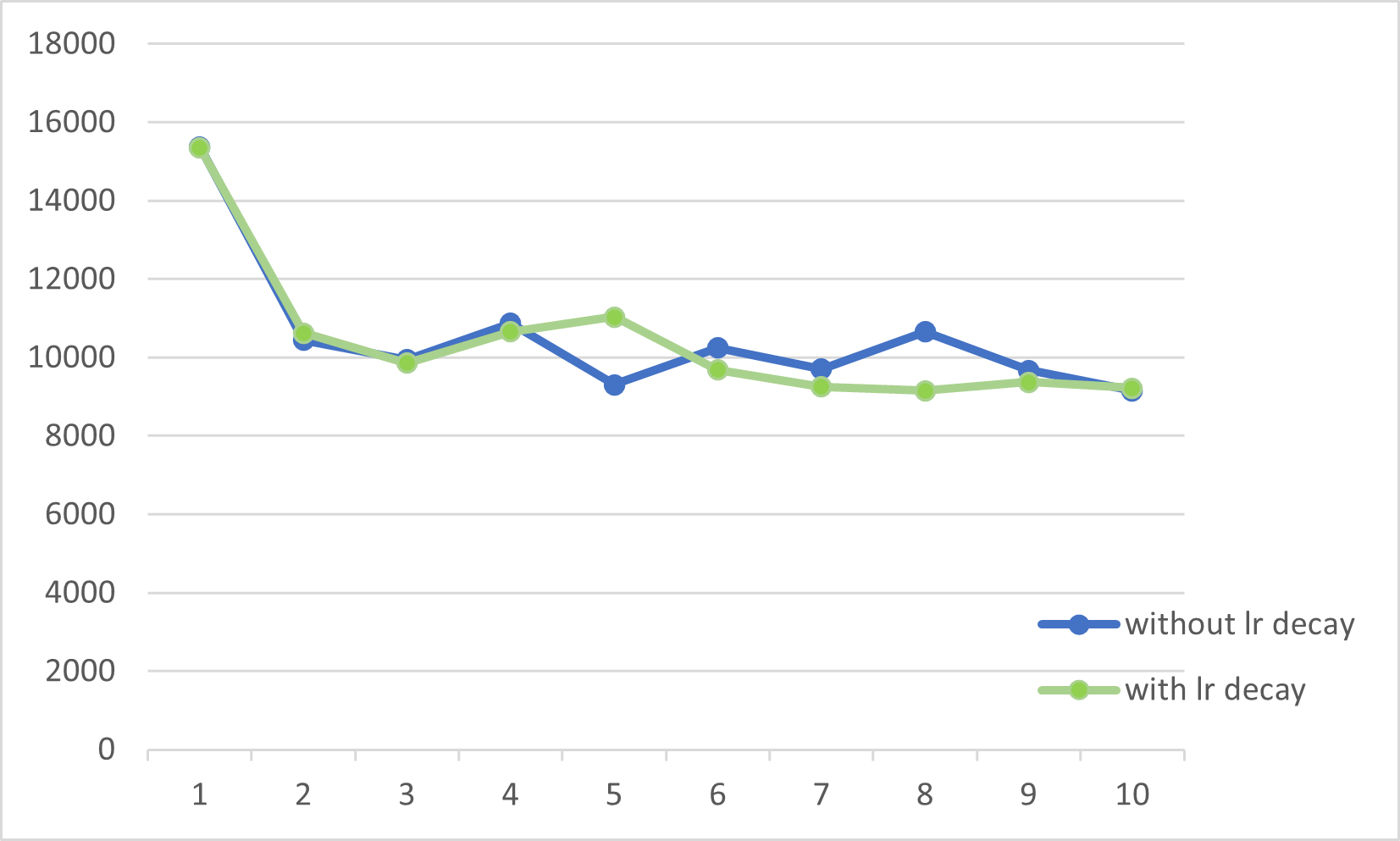}  
    \caption{Fluctuation of Hard Sample Count During Training}  
    \label{fig:fluctuation_of_hard_sample}  
\end{figure}

\subsection{Reward}

Our study adopted a simple 0-1 reward mechanism, which checks not only the correctness of the format but also the accuracy of the results. A reward of 1 will be assigned only if both the results are accurate and the format is correct; otherwise, the reward will be 0.

\[
R(x) = 
\begin{cases} 
1 & ResultCorrect \land FormatCorrect \\
0 & Otherwise,
\end{cases}
\tag{3}
\]
\textbf{Result checking.}
We conduct checks on the correctness of tool calls; specifically, the tool-call outputs generated by the model are parsed into a structured format (e.g., a Python dictionary) to facilitate precise verification. The model’s output must ensure the accuracy of both the tool names and the specific parameters—both of which need to exactly match the predefined ground truth (e.g., the tool name specified in the benchmark and the required argument values) to guarantee that the tool calls can be executed effectively without functional errors.

\textbf{Format checking.}
In training sessions utilizing Qwen’s official system prompts, we removed the format constraint on the <think></think> tags for the model. Concurrently, two key requirements were imposed: first, strict adherence to Qwen’s standardized output format; second, confinement of the model’s output exclusively within the <tool\_call></tool\_call> tags. By contrast, during the reconstruction training of Tool-N1, the format constraint on the <think></think> tags was retained without modification.

\section{Self-Exemplifying Thinking}
Think reasoning's effectiveness has been validated by numerous researchers.\citep{Wei2022ChainOT} The ``think" mechanism has been widely adopted in the training paradigm of the GRPO algorithm\citep{deepseekr1}. However, in these small-scale models, our guidance method fails to enable the ``think" mechanism to perform few-shot analysis. To address this issue, we aim to explore an approach that can effectively enhance the data utilization efficiency of both the ``think" mechanism and few-shot learning during the training process. Meanwhile, this approach should help the model break free from its reliance on few-shot guidance data, allowing it to autonomously generate few-shot examples even in scenarios where no pre-provided few-shot data is available. In this section, we will present detailed experiments and in-depth discussions around this objective.


\subsection{GRPO Modification for self-exemplifying}
Our goal requires the model to self-exemplify while further engaging in thinking. During the process of examples generation, it is critical to prevent the model from exhibiting reward hacking\citep{rewardhack}—a phenomenon in which the model obtains spurious rewards not by generating task-compliant. LLM in rule-based RL training might engage in undesirable generation behaviors. In this part, these behaviors include producing examples through trivial and meaningless modifications (e.g., replacing non-semantic stopwords without altering the core logic of the examples) or generating instances that are nearly redundant or repetitive (e.g., replicating existing example structures with only minimal adjustments to irrelevant details). Importantly, such reward hacking-related issues were observed in the initial phase of our experiments without modification. The distribution of the model will differ significantly from the original model distribution, and excessive constraints will lead to reduced exploration and limited performance. Drawing on the work of DAPO researchers, we have made appropriate modifications to the GRPO algorithm, further improving GRPO's performance.

\textbf{Clip Higher Policy.} 
Guided by few-shot examples, the entropy curve decreases more rapidly, and the early-stage exploration becomes more stable. We don't need to worry that the Clip higher strategy will easily cause the model training to collapse. The decoupling operation can increase the entropy in the later stage and encourage exploration. Since our goal is to enable the model to learn to think by generating its own examples, this strategy can make the model more bold when creating self-generated examples.

\textbf{Removing KL Divergence.}
The KL constraint appears to be of little necessity in our experimental setup. Our observations indicate that incorporating the KL constraint yields negligible benefits for performance improvement. Moreover, it introduces additional computational overhead, making the training process less efficient. Given these findings, removing the KL constraint not only avoids unnecessary computational burdens but also does not compromise the model's learning efficacy, thereby aligning with the goal of optimizing the training framework for both performance and efficiency.

Formally, given the historical context \( c_t \) and the set of currently accessible tools \( \mathcal{Z} \), the Qwen model generates a set of candidate responses \( \{ O^1, O^2, \dots, O^N \} \), where each \( O^n \in \mathcal{O} \). Here, \( \mathcal{O} \) denotes the space of possible output responses, and each response consists of two components: (1) textual reasoning tailored to tool-augmented problem-solving, and (2) an associated tool-related action \( a_n \). These candidate responses are evaluated by a reward function, producing a reward set \( \{ r_1, r_2, \dots, r_N \} \). We then optimize the policy \( \pi_\theta \) using GRPO, with the optimization objective formulated as:

\[
\begin{aligned}
\mathcal{L}_{\text{GRPO\ Modified}}(\theta) =& \mathbb{E}_{(c_t, \mathcal{Z})} \mathbb{E}_{O^i \sim \mathcal{O}} \Bigg[ \min \left( \rho_i A_i, \text{clip}(\rho_i, 1 - \epsilon_{lower}, 1 + \epsilon_{higher}) A_i \right)
 \Bigg], \text{where } \rho_i = \frac{\pi_\theta(O^i \mid c_t, \mathcal{Z})}{\pi_{\text{old}}(O^i \mid c_t, \mathcal{Z})}.
\end{aligned}
\tag{4}
\]

In this objective, \( \epsilon \) and \( \beta \) are tunable hyperparameters that balance the policy update range and the KL divergence constraint, respectively. \( A_i \) represents the relative advantage of the \( i \)-th response, which normalizes the raw reward \( r_i \) by the statistical properties of all candidate rewards in the batch:

\[
A_i = \frac{r_i - \text{mean}(\{ r_1, r_2, \dots, r_N \})}{\text{std}(\{ r_1, r_2, \dots, r_N \})},
\tag{5}
\]

where \( \text{mean} \) and \( \text{std} \) denote the mean and standard deviation of the rewards, ensuring that the advantage signal is scaled to guide policy updates effectively. In subsequent sections, we elaborate on key components of our approach: (1) the curation of tool-using data and its integration into the reinforcement learning pipeline, (2) the structured reasoning template (aligned with Qwen's prompt design) used during training, and (3) the reward modeling strategy customized for evaluating tool-augmented reasoning quality.

In practice, our modified framework delivers markedly enhanced performance. With dynamic sampling, the few-shot guided model in non-thinking mode substantially outperforms vanilla GRPO and approaches the performance of its thinking-mode counterpart. Furthermore, when empowered to autonomously generate few-shot examples \( s_n \in \mathcal{S} \), the thinking-mode model develops superior analytical capabilities for self-constructed contexts.

Subsequent sections detail experimental results of this framework, particularly the impact of removing KL constraints and decoupling clips on hard sample handling and overall performance.

\subsection{Reward}

We adopted a 0-1 reward mechanism. When the model's predicted answer is correct, if it successfully generates self-examples, we add a small additional reward of 0.01 to encourage the model to create few-shot examples on its own. Additionally, in modifying the system prompt, we added the <examples></examples> tags, requiring the model to provide examples for each tool within these tags. After completing the examples, the model needs to analyze by combining the examples with the user's question, and the analysis content should be placed within the  tags. During the experiment, it was found that the model would analyze the examples in the ``think'' section and break down the user's needs, and its performance was also better than the simple ``think'' mode of Tool-N1.
\[
R(x) = 
\begin{cases} 
1.01 & ResultCorrect \land FewshotsCorrect \land FormatCorrect, \\
1 & ResultCorrect \land FormatCorrect \\
0 & Otherwise
\end{cases}
\tag{6}
\]

\textbf{Result checking.}
We conduct checks on the correctness of tool calls. The tool-call output is parsed into a dictionary format, allowing for exact matching against the ground truth. This process involves two aspects: firstly, verifying whether the predicted tool name matches the ground truth, and secondly, confirming that all required arguments are correctly present. This strict matching criterion ensures that the model learns to generate functionally accurate and executable tool calls.

\textbf{Format checking.}
For the model's output formats, we experimented with Qwen's official function call output format and the output format provided by Tool-N1. For these distinct formats, we extracted the model's output results using regular expressions.
We conduct checks on the correctness of tool calls. The tool-call output is parsed into a dictionary format, allowing for exact matching against the ground truth. This process involves two aspects: firstly, verifying whether the predicted tool name matches the ground truth, and secondly, confirming that all required arguments are correctly present. This strict matching criterion ensures that the model learns to generate functionally accurate and executable tool calls.

\textbf{Few-shots checking.}
If the model generates a correct answer while simultaneously producing correct examples, we provide a very small additional reward.
We standardize the model to place its self-generated examples within the <examples></examples> tags. Specifically, we require the model to independently generate examples in a Json format,these examples must include tools, questions, and answers. To encourage the model to generate as many examples as possible, it is only eligible for the additional reward when it produces more than 3 distinct examples.In the absence of this small reward mechanism, the model tends to reiterate the answer within the <examples></examples> tags.We extracted each generated example from the examples set and verified the format correctness solely by checking whether the examples could be imported as Python dictionary via JSON format validation. Meanwhile, we imposed a requirement on the quantity of examples to ensure it meets the specified criteria. For the generated examples, no complex reward mechanisms were designed. Additionally, throughout the entire training process, the model consistently demonstrated a strong ability to draw inferences about other cases from one instance.


\begin{figure}[htbp]  
    \centering  
    \includegraphics[width=0.8\textwidth]{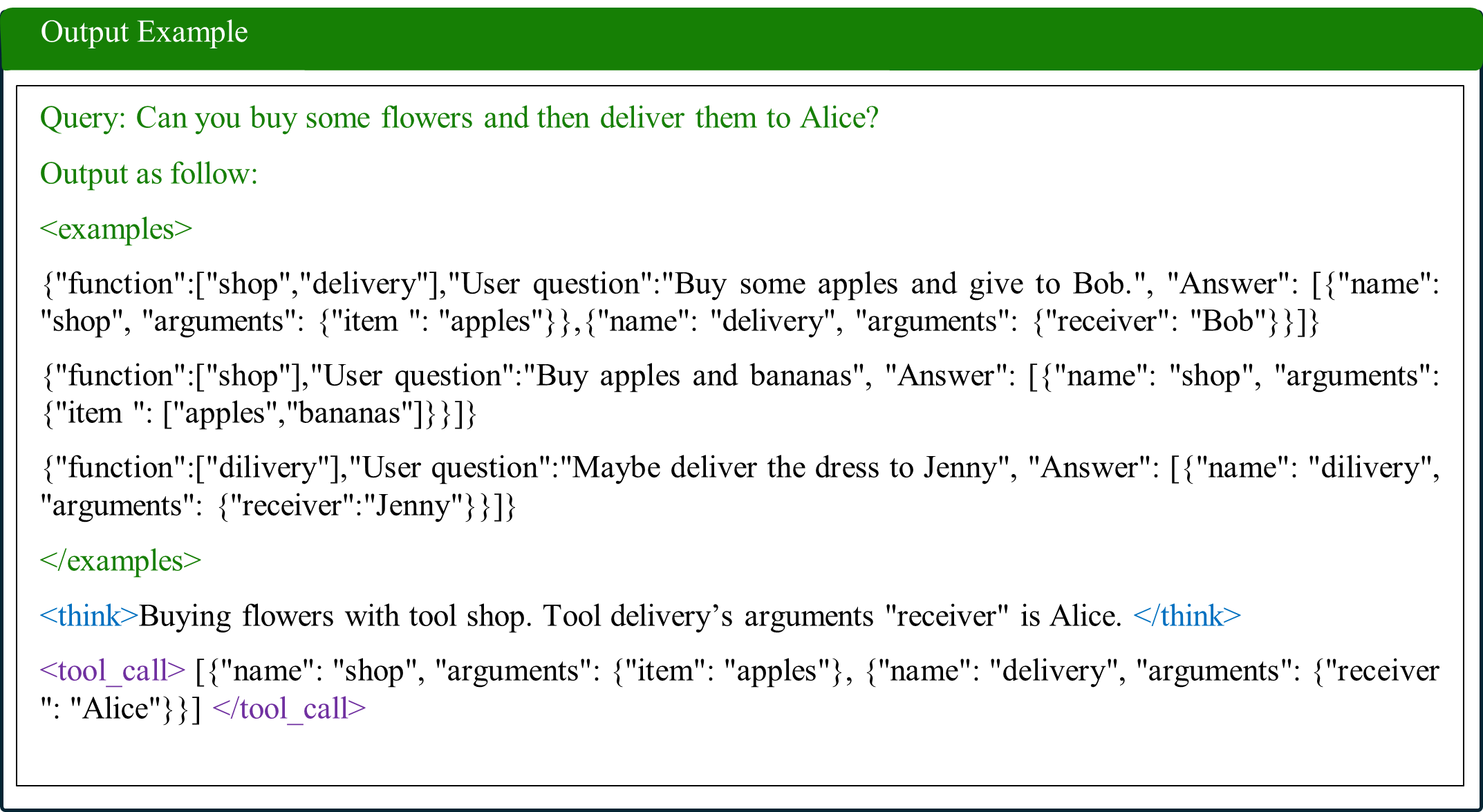}  
    \caption{Fully Correct Reward Results of few-shot Generated by the Model During Training}  
\end{figure}


\section{Experiments}

\subsection{Experiment details}
\textbf{Datasets} 
We used ToolACE and xLAM as the training data, while retaining the data composition adopted in the Tool-N1 work. Meanwhile, we constructed auxiliary data containing few-shot based on the original training data, which serves the dynamic hard sample sampling strategy. For the system prompts in the original data, we modified them respectively to Qwen's official system prompts and the system prompts designed for the self-exemplifying thinking mode. Those modifications were intended to compare the results of different experimental setups.

\textbf{Models}
We used Qwen2.5 and Qwen2 as the backbone models respectively. During the experiment, we found that Qwen2.5 seems to contain information from the original training data, which is reflected in the fact that the performance of Qwen2.5-1.5b-instruct is significantly higher than that of Qwen2-7b. Meanwhile, the number of hard samples of Qwen2-1.5b-Instruct on the training set is significantly greater than that of Qwen2.5-1.5-instruct.

\textbf{Benchmarks}
We mainly validate the single-turn tool-using scenarios, including the Berkeley Function Call Leaderboard (BFCL), APIBank, and ACEBench\citep{acebench}. We adopt a similar evaluation approach as the Tool-N1 work. For BFCL, we conduct evaluations on both the Non-live and Live subsets, corresponding to synthetic and real-world data. Each subset includes four categories: Simple, Multiple, Parallel, and Parallel Multiple. For APIBank, we find that the method of Tool-N1 seems to have no improvement at level-2, but our method achieves performance improvement at level-2. For ACEBench, we focus on two sub-categories within the Normal category: Atom and Single-turn.

\textbf{More Details:} We conducted our training with the VeRL\citep{sheng2025hybridflow} framework. All experiments were configured with 8 NVIDIA A800 80GB GPUs, and a batch size of 1024 was adopted. The temperature parameter was fixed at 0.7, while the KL divergence loss coefficient was set to 1 × $10^{-3}$ consistently across all experimental trials.

\subsection{Dynamic multi-round hard sampling}
Researchers of Tool-N1 noted that when employing the GRPO algorithm to maintain the model's reasoning (``think'') process, the improvement in the performance of small-scale models was quite limited with almost no improvement observed particularly on the Non-live subset of the BFCL benchmark. Similarly, in our research, when adopting the GRPO training paradigm alone, we also encountered issues where the training of small-scale models was highly unstable, often leading to training failures. However, under the few-shot guided training approach, the performance of small-scale models improved significantly. Meanwhile, for the 7B-parameter model, by directly utilizing Qwen's official system prompts for function calls (without adopting the ``think'' mechanism), we achieved performance that surpassed that of the paradigm with the ``think'' process enabled. More results are shown in Table \ref{tab:bfclperformance}.

APIBank is a tool-using benchmark with two modes: Call and Retrieve + Call. On the APIBank benchmark, our model significantly outperforms the training paradigm proposed by Tool-N1. Seen in Table \ref{tab:apibankacc}, this superiority is particularly evident on the Level-2, where it appears that the training method of Tool-N1 does not perform well.
 
\begin{table}[htbp!]
  \centering
  \footnotesize 
  \setlength{\tabcolsep}{1.5pt} 
  \begin{tabularx}{\textwidth}{l *{11}{>{\centering\arraybackslash}X}}
    \toprule
    \multirow{2}{*}{\textbf{Models}} & \multicolumn{4}{c}{\textbf{Non- Live}} & \multicolumn{4}{c}{\textbf{Live}} & \multicolumn{3}{c}{\textbf{Overall}} \\
    \cmidrule(lr){2-5} \cmidrule(lr){6-9} \cmidrule(lr){10-12}
    & Multiple & \makecell{Parallel\\Multiple} & Parallel & Simple & Multiple & \makecell{Parallel\\Multiple} & Parallel & Simple & Non-live ACC & Live ACC & Overall ACC \\
    \midrule
    ToolN1-7b & 0.94 & 0.91 & 0.925 & 0.955 & 0.7876 & 0.7083 & 0.75 & 0.81 & 0.937 & 0.8417 & 0.8525 \\

    Qwen2.5-7b-inst & 0.945 & 0.845 & 0.915 & 0.96 & 0.7493 & 0.7083 & 0.625 & 0.7674 & 0.925 & 0.7505 & 0.8248 \\
    Qwen2.5-7b-GRPO & 0.96 & 0.905 & 0.9 & 0.9525 & 0.7682 & 0.7917 & 0.875 & 0.818 & 0.934 & 0.8332 & 0.8452 \\
    Qwen2.5-7b-replace& 0.98 & 0.9 & 0.905 & 0.955 & 0.789 & 0.7083 & 0.625 & 0.837 & 0.939 & 0.8440 & \textbf{0.8561} \\
    Qwen2.5-7b-add & 0.975 & 0.9 & 0.925 & 0.955 & 0.7834 & 0.7083 & 0.625 & 0.798 & 0.942 & 0.8380 & 0.8506 \\
    \hline
    Qwen2-7b-inst & 0.88 & 0.75 & 0.755 & 0.8425 & 0.62 & 0.6667 & 0.375 & 0.5659 & 0.814 & 0.6893 & 0.6954 \\
    Qwen2-7b-GRPO & 0.945 & 0.855 & 0.9 & 0.9375 & 0.6895 & 0.7083 & 0.375 & 0.7752 & 0.915 & 0.7833 & 0.7929 \\
    Qwen2-7b-replace & 0.935 & 0.89 & 0.91 & 0.9325 & 0.7265 & 0.7083 & 0.5 & 0.7868 & 0.92 & 0.8037 & 0.8137 \\
    Qwen2-7b-add & 0.94 & 0.78 & 0.86 & 0.9325 & 0.7635 & 0.625 & 0.5 & 0.8062 & 0.889 & 0.8069 & \textbf{0.8184} \\
    \hline
    ToolN1-1.5b & - & - & - & - & - & - & - & - & 0.73 & 0.69 & 0.7083 \\
    Qwen2.5-1.5b-inst & 0.86 & 0.665 & 0.7 & 0.89 & 0.5926 & 0.4166 & 0.5625 & 0.7054 & 0.801 & 0.6857 & 0.6916 \\
    Qwen2.5-1.5b-replace & 0.91 & 0.775 & 0.785 & 0.8975 & 0.6952 & 0.5833 & 0.5625 & 0.81 & 0.853 & 0.7637 & 0.7729 \\
    Qwen2.5-1.5b-add & 0.92 & 0.83 & 0.86 & 0.9375 & 0.6818 & 0.5833 & 0.5 & 0.7287 & 0.897 & 0.7681 & \textbf{0.7762} \\
    \hline
    Qwen2-1.5b-inst & 0.79 & 0.405 & 0.465 & 0.795 & 0.4026 & 0.25 & 0.125 & 0.4884 & 0.65 & 0.5054 & 0.5138 \\
    Qwen2-1.5b-replace & 0.885 & 0.755 & 0.77 & 0.8775 & 0.5527 & 0.375 & 0.4375 & 0.67 & 0.833 & 0.6753 & \textbf{0.6822} \\
    Qwen2-1.5b-add & 0.89 & 0.67 & 0.725 & 0.86 & 0.5536 & 0.4583 & 0.375 & 0.6744 & 0.801 & 0.6641 & 0.6699 \\
    \bottomrule
  \end{tabularx}
\caption{Dynamic multi-round hard sampling's performance on BFCL benchmark.For ToolN1-1.5b, due to the GRPO training collapse issues we encountered during the training process, we have referenced approximate values from the data in the ToolN1. The calculation of ACC (Accuracy) is based on the number of correct samples divided by the total number of samples.``add" means few-shot guided samples are added in original training data under dynamic multi-round hard sampling policy, while ``replace" means replacing hard sample with few-shot guided sample }
\label{tab:bfclperformance}
\end{table}
\begin{table}[ht!]
    \centering
    \begin{minipage}{0.55\textwidth}
        \centering
        \begin{tabular}{lccc} 
            \hline 
            Model & Level-1 ACC & Level-2 ACC & ACC \\ 
            \hline
            
            ToolN1-7b & 81.2 & 58.209 & 77.8944  \\
            Qwen2.5-7b-inst & 76.692 & 68.657 & 75.5367 \\
            Qwen2.5-7b-GRPO & 66.165 & 41.794 & 62.6606 \\
            Qwen2.5-7b-replace & \textbf{82.957} & \textbf{74.627} & \textbf{81.7593}  \\
            Qwen2.5-7b-add & 79.449 & 71.642 & 78.3265 \\
            \hline
            Qwen2-7b-inst & 69.173 & 70.149 & 60.3133  \\
            Qwen2-7b-GRPO & 77.682 & \textbf{78.697} & 77.8279  \\
            Qwen2-7b-replace & 79.198 & 71.642 & 78.1116  \\
            Qwen2-7b-add & \textbf{81.704} & 71.642 & \textbf{80.257}  \\
            \hline
            Qwen2.5-1.5b-inst & 61.91 & 34.33 & 57.94 \\
            Qwen2.5-1.5b-replace & \textbf{76.942} & \textbf{68.657} & \textbf{75.7508}  \\
            Qwen2.5-1.5b-add & 73.183 & 65.672 & 72.103 \\
            \hline
            Qwen2-1.5b-inst & 52.13 & 49.254 & 51.7165  \\
            Qwen2-1.5b-replace & \textbf{68.17} & 52.239 & \textbf{65.8795}  \\
            Qwen2-1.5b-add & 64.411 & \textbf{55.224} & 63.0901  \\
            \hline        
        \end{tabular}
        \caption{Dynamic multi-round hard sampling's performance on APIBank}
        \label{tab:apibankacc}
    \end{minipage}
    \hfill 
    \begin{minipage}{0.35\textwidth}
        \centering
        \begin{tabular}{lc} 
            \hline 
            Model & ACC \\ 
            \hline
            
            ToolN1-7b & \textbf{80.4}  \\
            Qwen2.5-7b-inst & 64.8 \\
            Qwen2.5-7b-GRPO & 80.2 \\
            Qwen2.5-7b-replace & 78.4  \\
            Qwen2.5-7b-add & 74.0 \\
            \hline
            Qwen2-7b-inst & 63.0  \\
            Qwen2-7b-GRPO & 63.4  \\
            Qwen2-7b-replace & \textbf{67.0}  \\
            Qwen2-7b-add & 64.6  \\
            \hline
            Qwen2.5-1.5b-inst & 48.0 \\
            Qwen2.5-1.5b-replace & \textbf{62.6}  \\
            Qwen2.5-1.5b-add & 57.0 \\
            \hline
            Qwen2-1.5b-inst & 38.4  \\
            Qwen2-1.5b-replace & 46.8  \\
            Qwen2-1.5b-add & \textbf{50.2}  \\
            \hline        
        \end{tabular}
        \caption{Dynamic multi-round hard sampling's performance on ACEBench}
        \label{tab:acebenchacc}
    \end{minipage}
\end{table}

In the evaluations on the ACEBench leaderboard, the dynamic sampling strategy outperforms strategies trained directly with GRPO overall. However, the Qwen2.5-7b model, when adopting this dynamic sampling strategy, did not exhibit a significant advantage over the GRPO-based strategy. This phenomenon can be attributed to an inherent constraint in ACEBench’s evaluation protocol: the benchmark does not allow the use of training-phase system prompts. When the complexity of the test data increases and the model demonstrates a higher degree of fitting to the training data, the inability to reuse the system prompts from the training process adversely impacts model performance—ultimately preventing Qwen2.5-7b from achieving a noticeable performance breakthrough.

\textbf{Analysis:} Overall, within the dynamic multi-round hard sampling strategy, the ``replace" variant not only achieves superior performance compared to the ``add" variant but also exhibits significantly faster training speeds—attributed to its reduced dataset size.
In terms of model foundation, Qwen2.5 demonstrates notable advancements over Qwen2: it features an expanded pre-training dataset, has undergone specialized post-training for structured output tasks, and adopts a multi-stage reinforcement learning (RL) framework. \citep{Qwen25} Additionally, our analysis suggests that Qwen2.5 has likely been pre-trained on the original ToolACE and xLAM datasets; detailed insights supporting this observation can be found in the hard sample analysis table of the Qwen2.5.

\newpage
\subsection{Self-exemplifying Thinking}

We selected the model parameter that yielded the best evaluation results. Nevertheless, we observed that, despite performance fluctuations in some parameter configurations, all variants still slightly outperformed Tool-N1. This validates that the strong guidance of few-shot examples can also enhance training effectiveness during reinforcement learning when paired with a well-designed ``think" paradigm.

We also attempted a direct hard-sample removal approach; however, this method led to a performance that was much inferior to that of the hard-sample retention strategy. This underscores the value of preserving hard samples: some hard samples can be converted into non-hard samples during training, which aligns with the core objective of few-shot guidance, transforming hard samples into non-hard samples. 
Reasoning SFT enables the use of information from all samples; however, this direct application of SFT severely impairs the original capabilities of the model. In contrast, our proposed method can enhance the utilization efficiency of training data without compromising the model’s inherent performance.

Furthermore, we compared the performance of our model with Qwen3\citep{yang2025qwen3} and benchmarked it against other researchers' work, thereby further validating the cutting-edge nature of our approach.

\begin{table}[htbp!]
    \centering
    \begin{tabular}{lcccc} 
        \hline 
          strategy & Tool-N1 & self-exemplifying & self-exemplifying replace & Reasoning SFT \\ 
        \hline
        hard samples & 8527 & 8223 & 8048 & -\\
        BFCL ACC    & 0.8525 & \textbf{0.86} & 0.8498 & 0.7611 \\
        APIBank ACC    & 77.8944 & 80.22 & \textbf{80.24} & 74.23 \\
        ACEBench ACC    & 80.8 & \textbf{81.1} & 81.05 & 56.23  \\

        \hline
    \end{tabular}
    \caption{We adopt Qwen2.5-7b-Inst as the backbone model for this experiment, and meanwhile verify whether Dynamic Hard Sampling can further exert its performance in reducing the number of hard samples under this self-exemplifying thinking mode.} 
    \label{tab:self-exemplifying result} 
\end{table}
\begin{figure}[htbp]
    \centering
    \begin{subfigure}{0.45\textwidth}
        \centering
        \includegraphics[width=\linewidth]{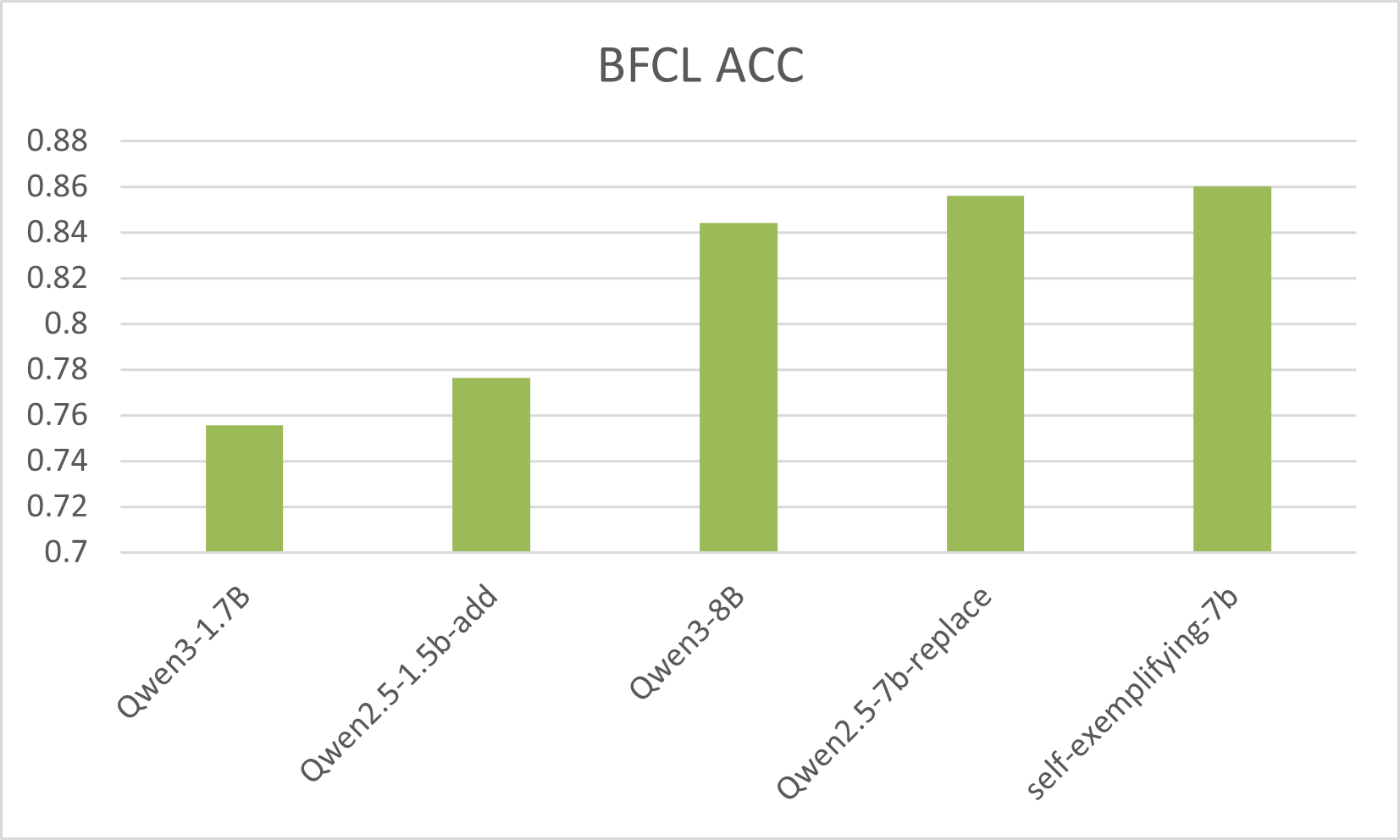}
        \caption{BFCL Accuracy Comparison}
        \label{fig:newestBFCL}
    \end{subfigure}
    \hfill  
    \begin{subfigure}{0.45\textwidth}
        \centering
        \includegraphics[width=\linewidth]{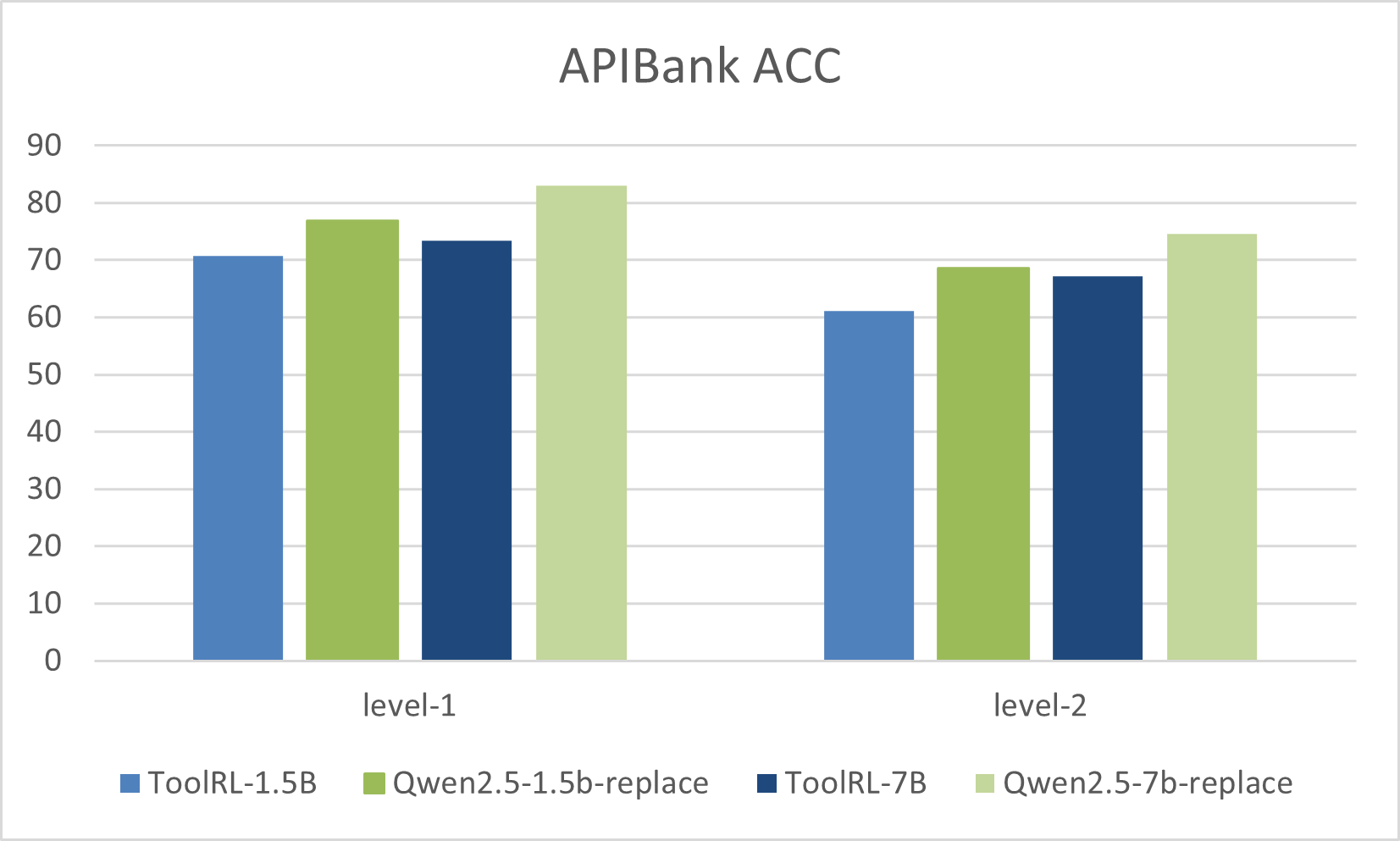}
        \caption{APIBank Accuracy Comparison}
        \label{fig:apibank}
    \end{subfigure}
    \caption{Accuracy Comparisons Across Different Model Configurations and Training Strategies. (a) Based on Data from BFCL List as of 2025-08-26 (b) Based on data from ToolRL:Reward is all you need\citep{toolrl}}
    \label{fig:accuracy_comparisons}
\end{figure}

We observed, the dynamic sampling strategy could further reduce the number of hard samples under the current thinking paradigm; however, this was accompanied by a decline in generalization performance. Regarding the performance improvements of Tool-N1 in the 14B and 7B model variants, part of the gains stemmed from the inherent capabilities of the models themselves, while another part originated from the dataset. In the later stages of training, the Tool-N1 reward curves of the 14B and 7B models tended to converge, with no significant differences in the reward values between them. This indicates that the performance enhancement of the 14B model was attributed more to its stronger intrinsic capacity, while the benefits derived from the dataset and the algorithm were relatively limited. Furthermore, the dataset contained a subset of contradictory samples. This interference resulted in minimal gains from the dataset, even as the number of hard samples decreased further. This also explains why the algorithm yielded more pronounced performance improvements for smaller-scale models.

\section{Conclusion}
We enhance Group Relative Policy Optimization (GRPO) for Large Language Models (LLMs) to tackle tool-using challenges in small-scale models, where ``hard samples" (with hard-to-obtain correct outputs) cause low data utilization, training instability, and collapse. Unlike existing methods that either rely on complex reward designs or remove hard samples, which lead to over-computation or data loss, we introduce two innovations. First, is Dynamic Multi-Round Hard Sampling, which dynamically swaps hard samples with quality few-shots during training, with exponential learning rate decay curbing oscillations. Second is Self-Exemplifying Thinking, a modified GRPO framework without KL divergence and with adjusted clip coefficients, plus a reward mechanism that incentivizes LLMs to autonomously generate, analyze few-shots, and optimize sampling endogenously. This work shows that targeted few-shot guidance and self-driven reasoning boost GRPO's efficiency for small-model tool-using, resolving critical data utilization, and stability bottlenecks.

\textbf{More} As this constitutes our preliminary draft, certain issues regarding structural details may be present. We will further refine and revise it in subsequent stages.

\newpage
\bibliographystyle{plainnat}
\bibliography{reference} 

\newpage
\appendix
\section{Benchmarks} 
\textbf{(1) BFCL}. BFCL evaluates the ability of Large Language Models (LLMs) to call functions accurately. Its data and leaderboard are constantly evolving, and this benchmark is widely used in tool learning research. The current version (v3) categorizes the data into three types: Live, Non-Live, and Multi-turn. This paper mainly focuses on the ``Live'' and ``Non-Live'' categories. Among them, ``Non-Live'' refers to data synthesized by researchers; ``Live'' data consists of queries contributed by real users. 

\textbf{(2) ACEBench}. ACEBench is designed to evaluate tool-use capabilities with fine-grained categorization, which can be divided into three primary categories: Normal, Special, and Agent. In this study, we focus on two sub-categories within the ``Normal'' category: Atom and Single-turn. ``Atom'' cases consist of API calls that utilize specific parameter types; the ``Single-turn'' subset encompasses both sequential and parallel tool-using scenarios. Different from the other two benchmarks, ACEBench does not use the system prompts adopted during training, which may lead to a certain deviation in evaluation and increase the impact of system prompts to a certain extent. 

\textbf{(3) APIBank}. APIBank is a tool-using benchmark with two modes: Call and Retrieve + Call. The model needs to interpret user intents from the dialogue and execute the corresponding local Python tools. The specific data is divided into three types of evaluation data: level-1, level-2, and level-3. Our tests mainly use the data of level-1 and level-2 for our evaluation experiments.

\section{More Implementation Details}  

To validate the effectiveness of Dynamic Multi-Round Hard Sampling, we compared its training results with those obtained under the training mode that uses Qwen’s official system prompts (without the ``think" emphasis). The training parameters for GRPO and Dynamic Multi-Round Hard Sampling are provided in the table below. The same set of parameters was adopted for both the 1.5B and 7B model variants.
\begin{table}[htbp!]
    \centering
    \begin{tabular}{llll} 
        \hline 
         Hyperparameter  & Value & Hyperparameter & value \\ 
        \hline
        Batch Size & 1024 & Learning Rate & 1e-6 \\
        Epoch Number    & 10 & Entropy Coefficient & 0 \\
        KL Coefficient  & 1e-3 & Rollout Number & 5 \\
        Temperature    & 0.7 & Max Response Length & 8192 \\
        clip coefficient & 0.2\\

        \hline
    \end{tabular}
    \caption{Training Parameters of GRPO and Tool-N1} 
    \label{tab:Training Parameters of GRPO and Tool-N1} 
\end{table}

\begin{table}[htbp!]
    \centering
    \begin{tabular}{llll} 
        \hline 
         Hyperparameter  & Value & Hyperparameter & value \\ 
        \hline
        Batch Size & 1024 & Learning Rate & 1e-6 \\
        Epoch Number    & 10 & Entropy Coefficient & 0 \\
        KL Coefficient  & 1e-3 & Rollout Number & 5 \\
        Temperature    & 0.7 & Max Response Length & 8192 \\
        Learning Rate Decay    & 0.8 & Clip Coefficient & 0.2  \\
        \hline
    \end{tabular}
    \caption{Training Parameters of Dynamic multi-round hard sampling strategy} 
    \label{tab:Training_Parameters_of_Dynamic_multi-round_hard_sampling_policy} 
\end{table}

To validate the effectiveness of the Self-exemplifying Thinking mindset, we compared the experimental results of Self-exemplifying Thinking with those of Tool-N1 in our experiments. The specific experimental parameters are provided in the table below. Practically, we directly adopted Tool-N1’s prompts and reward mechanism, while incorporating the dynamic multi-round hard sampling strategy. However, the results obtained under this setup were completely consistent with those of Tool-N1. This indicates that the ``think" (i.e., Self-exemplifying Thinking) element was directly overlooked, and few-shot examples exerted a dominant influence on the experimental outcomes.

\begin{table}[htbp!]
    \centering
    \begin{tabular}{llll} 
        \hline 
         Hyperparameter  & Value & Hyperparameter & value \\ 
        \hline
        Batch Size & 1024 & Learning Rate & 1e-6 \\
        Epoch Number    & 10 & Entropy Coefficient & 0 \\
        KL Coefficient  & Removed & Rollout Number & 5 \\
        Temperature    & 0.7 & Max Response Length & 8192 \\
        Clip Higher Coefficient    & 0.26 & Clip Lower Coefficient & 0.2  \\
        Learning Rate Decay    & 0.8 &  &  \\
        \hline
    \end{tabular}
    \caption{Training Parameters of self-exemplifying policy and self-exemplifying replace} 
    \label{tab:student_scores} 
\end{table}

\newpage
\section{More Curves}  
From the reward curves, when using the system prompt without the ``think" component, the overall trend of the reward curve is consistent when comparing the dynamic hard sampling strategy with the standard GRPO training. Meanwhile, it is evident that in the later stages of training, the GRPO reward curve guided by few-shot slightly surpasses that of the vanilla GRPO. Additionally, the reward curve of the Self-exemplifying (or Self-Amplifying, depending on your intended meaning) approach is significantly higher than that of the Tool-N1 approach—and this discrepancy is clearly not merely caused by minor numerical differences in additional rewards. In fact, the magnitude of the minor additional rewards can be set even smaller to better isolate and observe the actual impact of the training strategy itself. Furthermore, the model achieves notably higher rewards on Guidance data, which indicates that Guidance data is more conducive to generating correct results compared to raw data.

\begin{figure}[htbp!]  
    \centering  
    \begin{subfigure}[b]{0.31\textwidth}
        \centering
        \includegraphics[width=\textwidth, keepaspectratio]{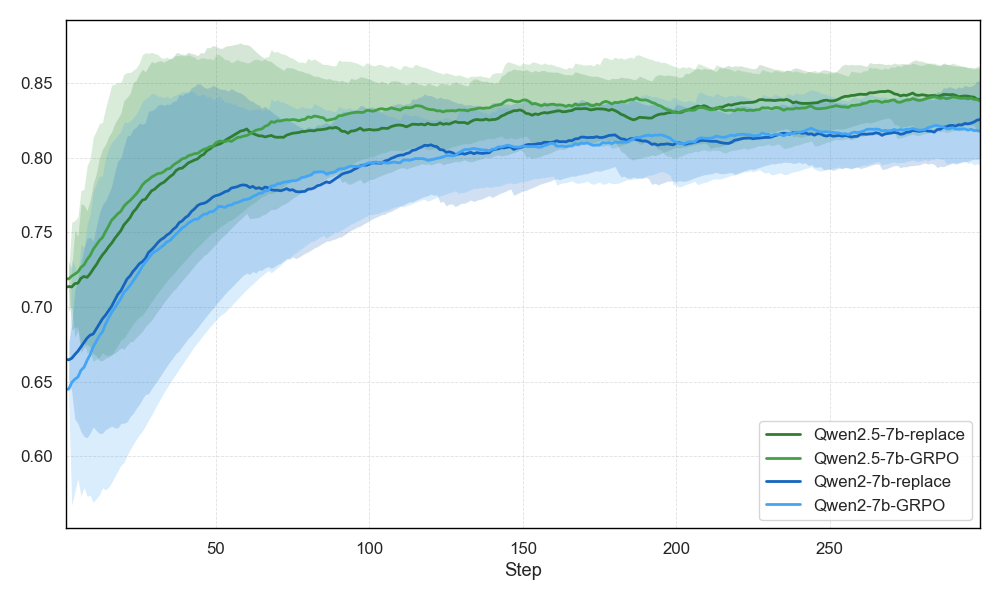}
        \caption{GRPO and GRPO-replace}  
        \label{fig:rew1}
    \end{subfigure}
    \hspace{0.02\textwidth}  
    \begin{subfigure}[b]{0.31\textwidth}
        \centering
        \includegraphics[width=\textwidth, keepaspectratio]{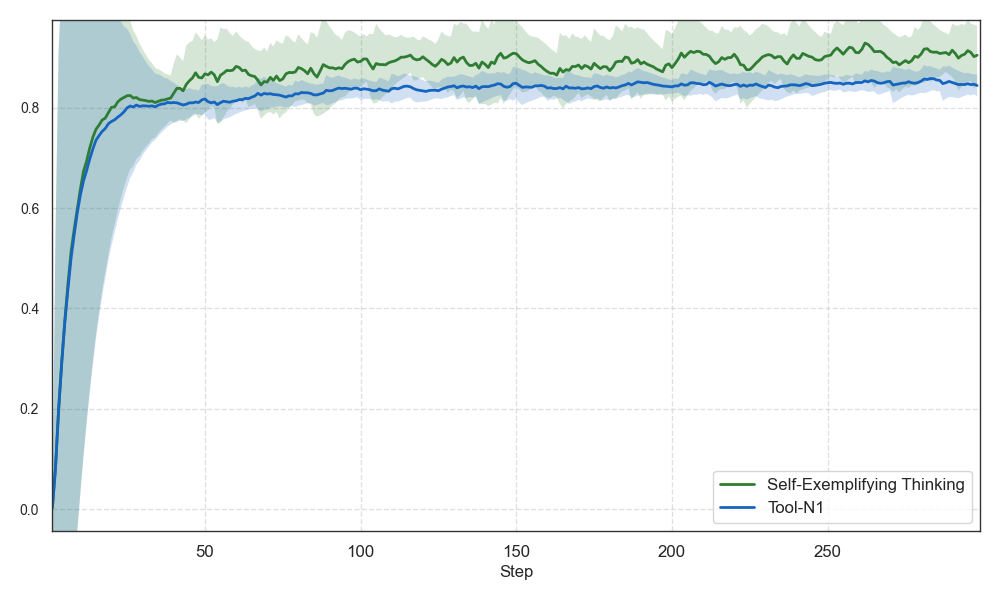}
        \caption{Self-exemplifying and Tool-N1}  
        \label{fig:rew2}
    \end{subfigure}
    \hspace{0.02\textwidth}  
    \begin{subfigure}[b]{0.31\textwidth}
        \centering
        \includegraphics[width=\textwidth, keepaspectratio]{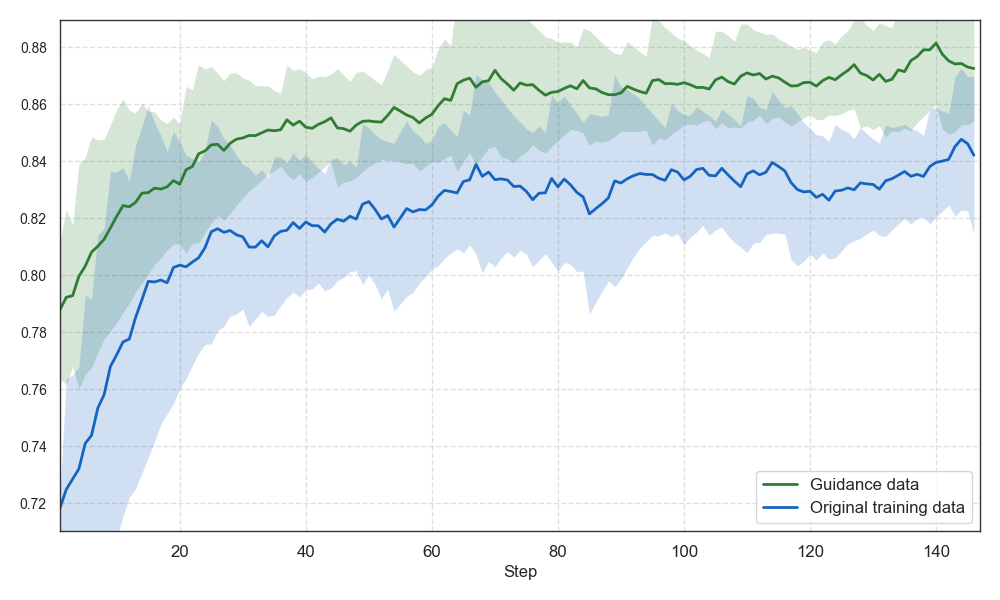}
        \caption{Guidance data and raw data}  
        \label{fig:rew3}
    \end{subfigure}
    \caption{Comparison of Three Reward Curves}
    \label{fig:rew_all}
\end{figure}

\section{Prompts}

\begin{figure}[htbp]  
    \centering  
    \includegraphics[width=0.8\textwidth]{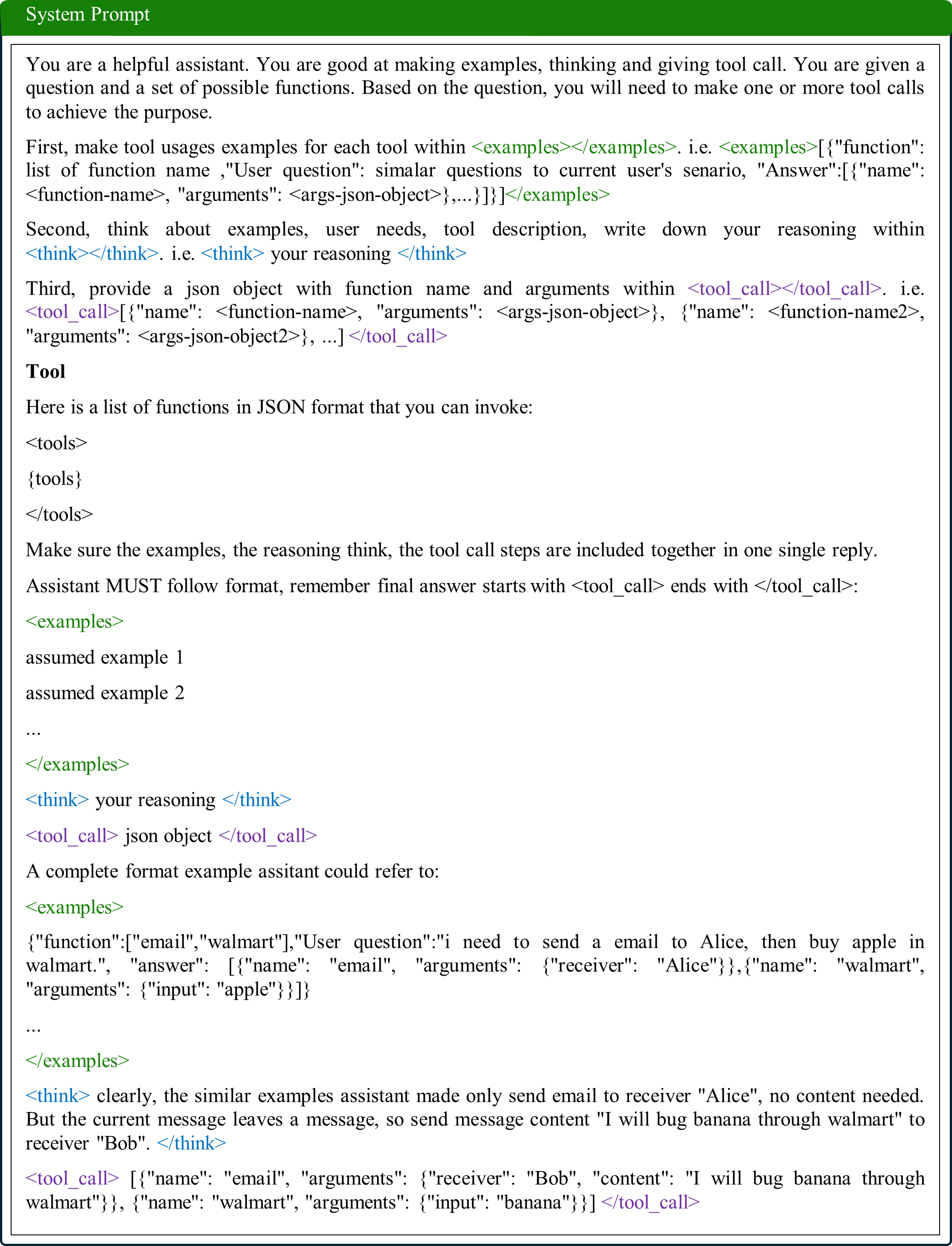}  
    \caption{the prompt template for the self-exemplifying thinking mode}  
    \label{fig:the prompt template for the self-exemplifying thinking mode}  
\end{figure}

In the validation of the dynamic hard sampling strategy, we did not employ a complex system prompt, but instead directly adopted the official system prompt provided by Qwen. Similarly, the original model was evaluated across multiple leaderboards using the same official Qwen system prompt, with the exception of ACEBench, for which only the benchmark-specific system prompt was used.
When further designing the training for the self-thinking mode, we designed a self-thinking prompt to guide the model to generate few-shot by self-exemplification and analyze few-shot examples during the thinking process(shown in Figure \ref{fig:the prompt template for the self-exemplifying thinking mode}). Our methodological design is predicated on a key prerequisite: the base model, when operating under the designated system prompt, must be capable of generating at least one correct response within its sampling rollouts. This ensures the availability of a positive learning signal for policy optimization. Our experimental validation confirms that this ``Self-Exemplifying Thinking" paradigm is highly compatible with our dynamic multi-round hard sampling strategy. The synergy between the structured reasoning facilitated by ``think" and the targeted guidance provided by few-shots collectively drives significant performance enhancement.

\begin{figure}[htbp]  
    \centering  
    \includegraphics[width=0.8\textwidth]{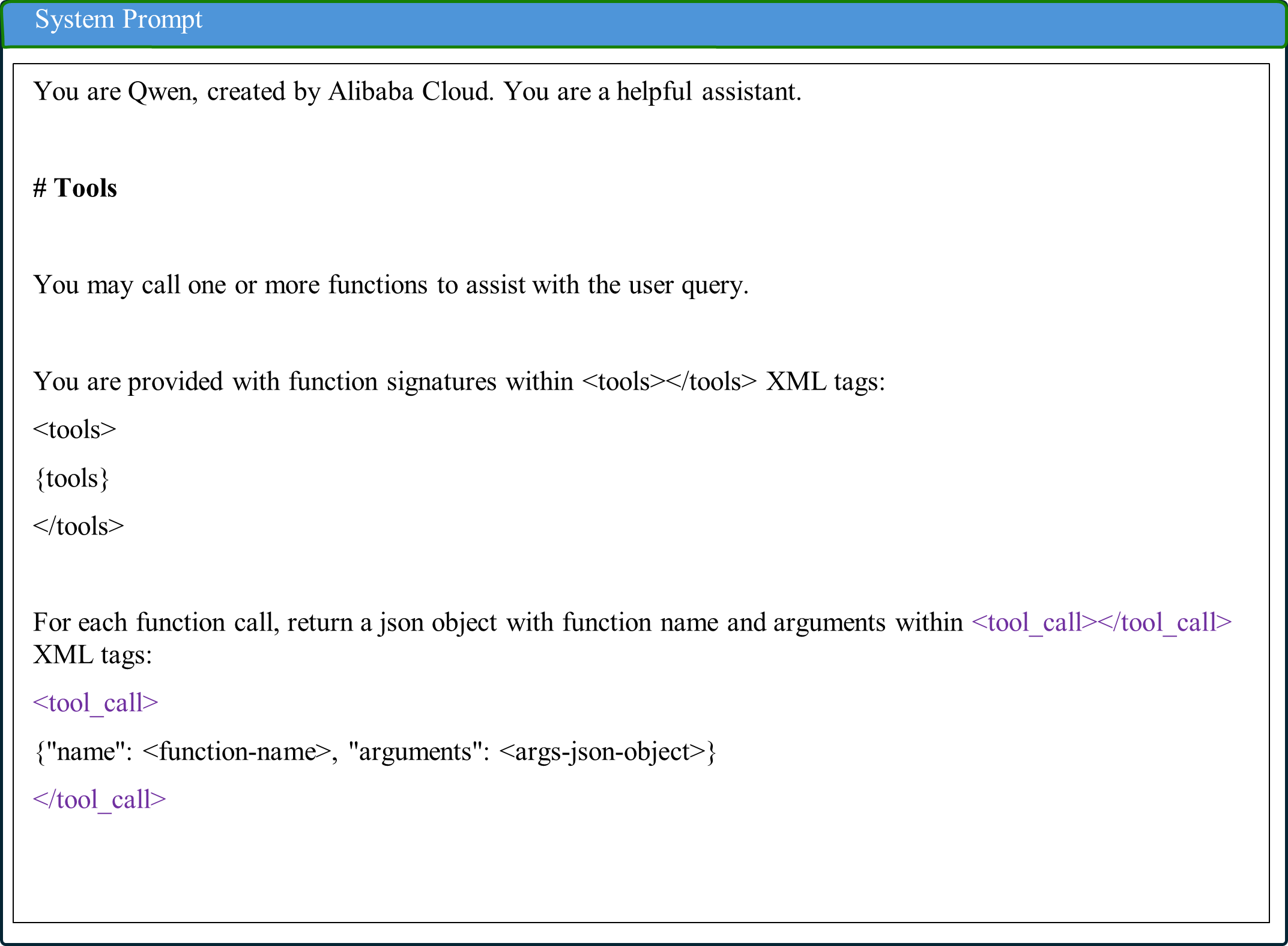}  
    \caption{Qwen prompt template }  
    \label{fig:Qwen prompt template}  
\end{figure}

\begin{figure}[htbp]  
    \centering  
    \includegraphics[width=0.8\textwidth]{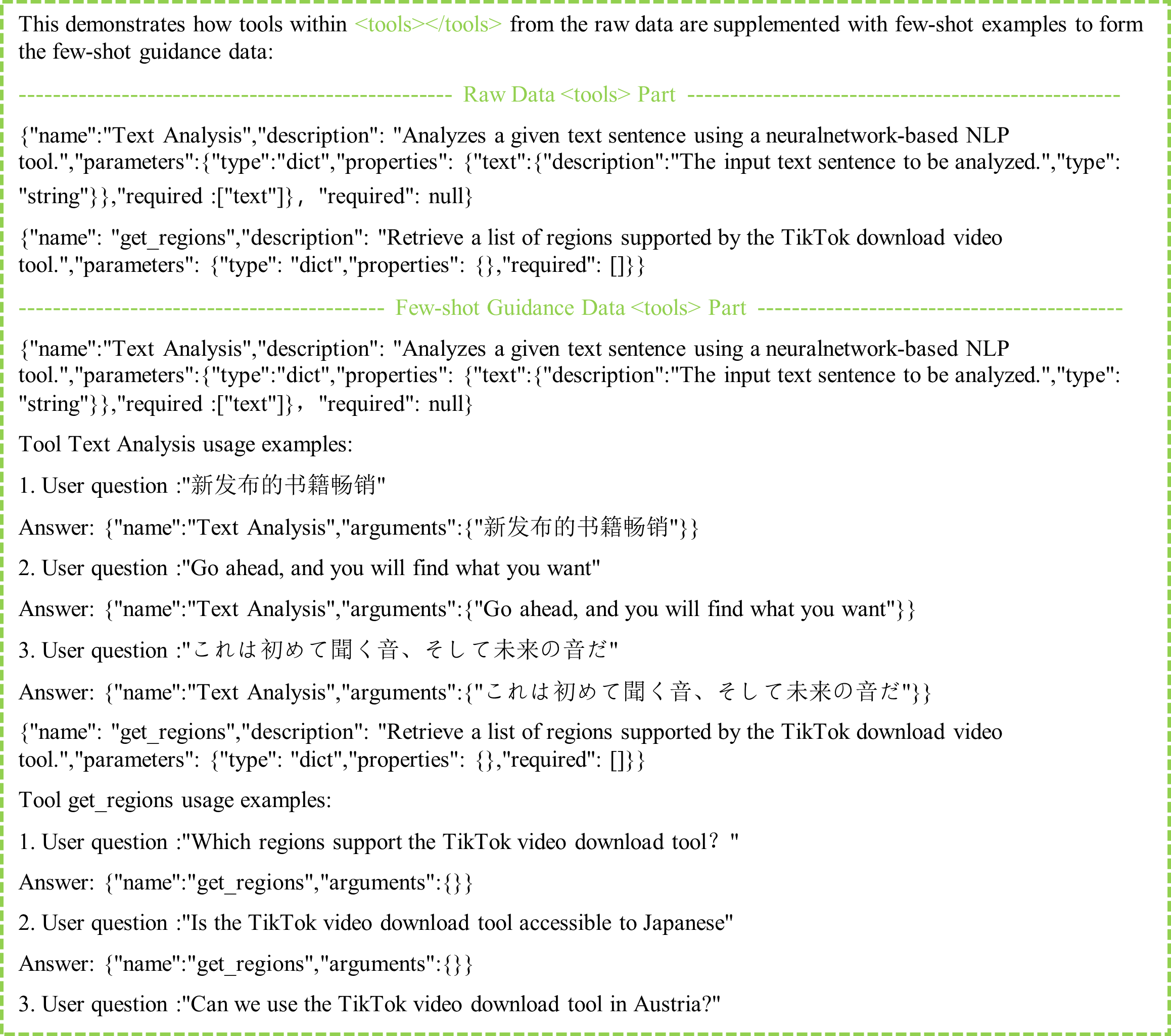}  
    \caption{Schematic of Few-shot Example Augmentation Approach for Raw Data}  
    \label{fig:Example Augmentation Approach}  
\end{figure}

\end{document}